\documentclass[5p,times,procedia]{elsarticle}
\usepackage{amsmath} 
\usepackage{amssymb}  

\usepackage{graphicx} 
\graphicspath{{images/}} 
\usepackage{subcaption}
\usepackage[export]{adjustbox}
\usepackage{caption}

\DeclareMathOperator*{\argminA}{arg\,min}
\usepackage{ecrc}

\volume{00}

\firstpage{1}
\journalname{Procedia CIRP}

\runauth{J. Elstner, R. Schönhof et. al.}

\jid{trpro}

\usepackage{amssymb}

\usepackage[figuresright]{rotating}

\usepackage[bookmarks=false]{hyperref}
    \hypersetup{colorlinks,
      linkcolor=blue,
      citecolor=blue,
      urlcolor=blue}

\begin{document}
\begin{frontmatter}



\dochead{33rd CIRP Design Conference}%

\title{Optimizing CAD Models with Latent Space Manipulation}


\author[a]{Jannes Elstner \textsuperscript{*,}} 
\author[a]{Raoul Schönhof \textsuperscript{*,**,}}
\author[b]{Steffen Tauber} 
\author[c]{Marco F. Huber}

\address[a]{Fraunhofer Institute for Manufacturing Engineering and Automation IPA, Nobelstraße 12, 70569 Stuttgart, Germany}
\address[b]{evia solution GmbH, Am Wallgraben 100, 70565 Stuttgart, Germany}
\address[c]{Institute of Industrial Manufacturing and Management IFF, University of Stuttgart, Allmandring 35, 70569 Stuttgart, Germany}

\aucores{* Equally contributing; ** Corresponding author. {\it E-mail address:} raoul.schoenhof@ipa-extern.fraunhofer.de}

\begin{abstract}
When it comes to the optimization of CAD models in the automation domain, neural networks currently play only a minor role. Optimizing abstract features such as automation capability is challenging, since they can be very difficult to simulate, are too complex for rule-based systems, and also have little to no data available for machine-learning methods. On the other hand, image manipulation methods that can manipulate abstract features in images such as StyleCLIP have seen much success. They rely on the latent space of pretrained generative adversarial networks, and could therefore also make use of the vast amount of unlabeled CAD data. In this paper, we show that such an approach is also suitable for optimizing abstract automation-related features of CAD parts. We achieved this by extending StyleCLIP to work with CAD models in the form of voxel models, which includes using a 3D StyleGAN and a custom classifier. Finally, we demonstrate the ability of our system for the optimiziation of automation-related features by optimizing the grabability of various CAD models.
\end{abstract}


\end{frontmatter}



\section{Introduction}

In the domain of manufacturing, assembly processes belong to the central aspects of costs and value creation \cite{Spur_86}. Thus, steady redaction of assembly cost is an important goal. One possibility to reduce costs is increasing the level of automation in the assembly processes. This aspect becomes even more important since labor becomes increasingly costly and skilled workers are rare. Unfortunately, whether or not a product can be assembled automatically depends on many parameters, like the material, surface properties, packaging, size, or the shape of the part. However, the influence of the part's shape is often especially critical for numerous aspects of the assembly process. For example: Parts may need to be separated from others parts (bulk goods), or get grasped by a robot. They also need to be designed to join other parts easily.
 \\ \\
Addressing these aspects is necessary to enable to be feasible assembly process. This requires considering the features and limitations of the automated assembly process in the product design. However, designing products and optimizing them for automation-ready-assembly requires substantial knowledge from the production side, and a system that is able to optimize CAD models towards better automation capability. Such a system does not exist yet. \\

With the NeuroCAD pipeline, we have already demonstrated, that several aspects like learning, the assessment, analysis and locating of automation-related features in CAD parts is possible \cite{Schoenhof_NN, Schoenhof_xAI, Schoenhof_AE}. \\

Within this work, we present a first approach towards optimizing CAD parts with respect to abstract, automation-related features. Our approach is based on insights from image-to-image translation, particularly from image manipulation methods that make use of the latent space of generative adversarial networks, such as StyleCLIP. Section 2 of this paper briefly describes the state of the art of our NeuroCAD pipeline, StyleGAN, and StyleCLIP approach\cite{Schoenhof_NN, Schoenhof_xAI, Schoenhof_AE, stylegan1, stylegan2, styleclip}. In section 3, we describe the architecture of our system for optimizing abstract features. Finally, we demonstrate the capability of our system in Section 4, by optimizing the grabability of components using a small dataset we built. Section 5 and 6 summarizes our work and provides an outlook for the future. 

\section{Related Work}
\subsection{CAD Optimization}

When it comes to the optimization of 3D topologies with neural networks, there are many approaches to a large variety of problem domains. From mechanical features, like stress, deformation, heat transfer or flow mechanics. Often these approaches replace optimization methods like evolutionary algorithms or search strategies. Especially in the domain of finite element method had been mentionable improvements, e.g., \cite{applmech1020007, Sumudu2021}.
However, usually an 3D shape optimization algorithm only tackles tasks, which are described as an analytical equation or a numerical simulation. Problems which can not be practically simulated also cannot be optimized. An optimization of 3D shapes, based on learned features does not exist yet. 
 
\subsection{Latent Space Manipulation}
In the last few years, many works explored image manipulation based on the latent space of a pretrained generative adversarial network (GAN) \cite{interfacegan, styleclip, age_manipulation, encoding_in_style}.
For latent space manipulation methods, the images are first embedded into the GAN's latent space in a process called GAN \textit{inversion}. The obtained latent code is then manipulated in the latent space, e.g. by traversing along directions that correspond to attributes such as the spatial layout of scenes \cite{lighting_scenes}.\\
StyleGAN \cite{stylegan1, stylegan2} is the most commonly used GAN for latent space manipulation because of its disentangled latent spaces, state-of-the-art image quality, and architecture that enables scale-specific manipulations.
\subsection{StyleCLIP}
StyleCLIP \cite{styleclip} is a method to manipulate images using text prompts. For this, StyleCLIP relies on a pretrained StyleGAN generator, and a pretrained Contrastive Language-Image Pre-training (CLIP) \cite{clip} model to guide the manipulation by assessing how well a text prompt fits to a given image. \\
StyleCLIP uses three different methods for the manipulation of images: \textit{latent optimization}, optimization using a \textit{latent mapper}, and \textit{global directions}. We base our CAD optimization on the former two methods. The global directions method seems unsuited, since it is based on the assumption that all image manipulations for an attribute of interest share the same step in latent space. This is too restricting for CAD optimization, since optimizations may have high specificity for a particular component.
\subsubsection{Latent Optimization}
The image manipulation is modeled as an optimization problem and solved by directly optimizing the latent code.
Specifically, given a source latent code $w_s$ out of StyleGAN's extended latent space $ \mathcal{W}+$, and a text prompt $t$ as directive, the image manipulation is modelled as the following optimization problem:
\begin{equation}
	\argminA_{w \in \mathcal{W}+} D_{CLIP}(G(w), t) + \lambda_{L2}\lvert\lvert w - w_s \rvert \rvert_2 + \lambda_{ID}\mathcal{L}_{ID}(w),
\label{equ:opt}
\end{equation}
where $D_{CLIP}$ is a function that computes the cosine distance between the CLIP embeddings of its two arguments \cite{styleclip}. The similarity between the manipulated image and the source image is controlled by the $L_2$ distance in latent space. Furthermore, an identity loss is added when manipulating human faces to preserve features establishing the person's identity.
\subsubsection{Latent Mapper}
The latent mapper is a mapping network that is trained for a specific text prompt $t$ and infers a manipulation step in latent space given the source latent code of an image. 
Given a dataset of source latent codes of images, the latent mapper $M_{t}$ is trained to minimize the following loss:
\begin{equation}
    \begin{aligned}
    	\mathcal{L}(w)=\:&D_{CLIP}(G(w+M_{t}(w)), t) \\
    	&+ \lambda_{L2}\lvert\lvert M_{t}(w) \rvert \rvert_2 \\
    	&+ \lambda_{ID}\mathcal{L}_{ID}(w).
    \end{aligned}
\label{equ:mapper_loss}
\end{equation}
Here, the optimized latent code is obtained by adding the inferred manipulation step $M_{t}(w)$ to the latent code $w$. The similarity between the manipulated image and the source image is controlled by the $L_2$ norm of the manipulation step.

\begin{figure*}[t!]
    \centering
    \includegraphics[width=\linewidth]{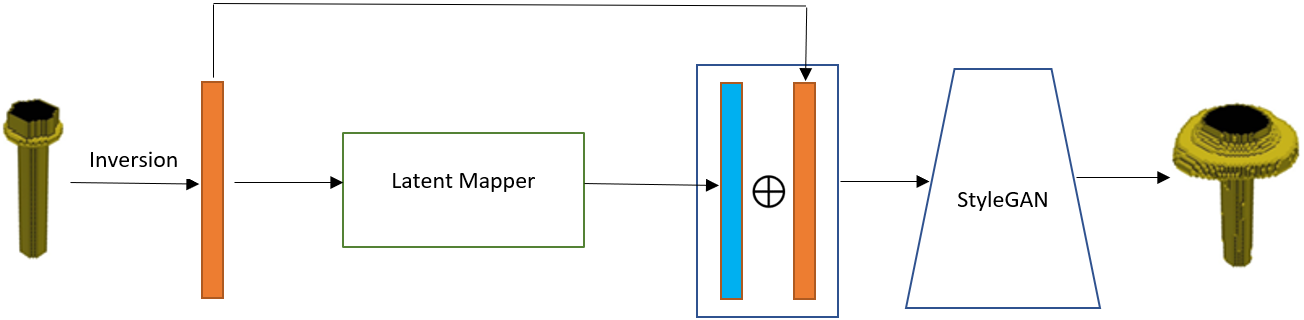}
    \caption{Grabability optimization using the latent mapper. The original component is inverted into a source latent code (orange), which is fed to the latent mapper. The mapper generates a vector (blue) that is added to the source latent code to yield the optimized latent code. The StyleGAN generates the optimized component from the optimized latent code. During training, the optimized component is compared to the original component using the comparator. Inspired by \cite{styleclip}.}
    \label{fig:latent_mapper}
\end{figure*}
\section{Methods}
We base our CAD optimization on StyleCLIP, utilizing both its latent optimization method and latent mapper method. 
StyleCLIP relies on StyleGAN for its disentangled latent space and CLIP for its vision-language representations. To use StyleCLIP for the optimization of CAD models, the StyleGAN needs to be able to encode CAD models, and CLIP has to be replaced with a loss network that can assess the optimization objective. For the latter we use a comparator network, which compares the original component with the optimized component in terms of their degree of optimization. Note that compared to StyleCLIP, our optimization is not text-driven and we do not use any natural language processing in our system.\\
We represent our CAD models in a voxel-based representation, since voxel grids are the 3D analogon of 2D images, which most latent space manipulation methods operate on. Specifically, we rasterize our data into binary grids of 64x64x64 resolution to capture as many details as possible while remaining in the computational budget. The dilemma of choosing the right resolution is typical for voxel models and it has been discussed in our previous work \cite{Schoenhof_NN}.

\subsection{StyleGAN} 
To train StyleGAN on 3D voxel models, we use a setup similar to \cite{stylegan3d}, replacing all 2D operations with their respective 3D equivalents and  adding a tanh activation function at the end of the generator. Furthermore, we remove the noise inputs in the generator since they model stochastic variation \cite{stylegan2}, which CAD models typically do not have. Lastly, to aid training stability and convergence we use class-conditioning as described in \cite{ada}.\\
Given the high GPU memory requirements of 3D deep learning, we significantly reduced the number of StyleGAN's trainable parameters. The original StyleGAN2 generator and discriminator each have about 30M trainable parameters, which we reduced to about 2.1M parameters for each. 
\\
For the evaluation of GANs, the Fréchet Inception Distance (FID) \cite{fid} is the most commonly used metric. The FID relies on a InceptionV3 \cite{inceptionv3} network trained on 2D natural images, and can therefore not directly be used for the evaluation of 3D GANs. We follow \cite{stylegan3d} by instead calculating the FID scores on the middle slices of axial, coronal and sagittal planes of the synthesized 3D voxel models.\\
After training and evaluation of the StyleGAN, we use a simplified version of the inversion method proposed in \cite{stylegan2} to embed components in the generator's latent space.
\subsection{Comparator}
The CLIP model needs to be replaced with a loss network that can assess the optimization objective. Typically, the loss network is either a classifier or a regressor, but we instead opt to use a comparator network \cite{comparator}, which compares a pair of CAD models based on their degree of optimization. \\
The main advantage of using a comparator is that building the dataset is significantly easier, especially in the use case we are most interested in: the optimization of automation properties. Compared to scoring CAD models on automation properties, it is easier to determine which of two given models scores better. 
\\
On the other hand, a comparator can only answer which of two given models is better, but not by how much. Also, specifying a concrete optimization goal, e.g. a desired target age as in \cite{age_manipulation}, is not possible. When it comes to optimization, this does not pose a problem, since targets are typically hard to establish or justify. Using the optimization of production costs as an example, any optimal price chosen as a target is inherently arbitrary. In practice, the starting point for the optimization would be to want the CAD model to just cost less. \\
The comparator architecture is based on the 2-channel network proposed by \cite{comparator}. When comparing two CAD models, they are concatenated along their channel-axis and fed to a convolutional network as a single two-channel image.\\
Given voxel models $V_1$ and $V_2$, the comparator is trained to predict which of the two is more optimized. This can be achieved by training the comparator to minimize the following loss:
\begin{equation}
	\mathcal{L}(V_1 , V_2, y_{true}) = H(y_{true}, C(V_1 , V_2)),
    \label{equ:comp_loss}
\end{equation}
where $C$ is the comparator, $y_{true}$ the label, and $H$ computes the binary cross-entropy between its arguments. The optimization comparison is hence treated as a binary
classification problem with the two classes ”$V_1$ is more optimized” and ”$V_2$ is more optimized / $V_1$ is less optimized".

\subsection{Optimization Methods}
Our optimization methods operate in StyleGAN's intermediate latent space $\mathcal{W}$ space instead of the original extended latent space $\mathcal{W+}$, since using $\mathcal{W}$ led to higher perceptual quality and more consistent results. This also means that for our latent mapper, we use StyleCLIP's setup with a single mapping network instead of three mapping networks for different feature levels.\\
For the loss functions, we replace the CLIP loss with an optimization loss that uses the comparator to compare the source component with the optimized component. 
The similarity between the optimized component and the original component is still controlled by the $L_2$ norm of the inferred manipulation step in latent space, but with the addition of an $L_2$ loss in data space since the mapping from latent space to data space is not always smooth. We also remove the identity loss, because CAD models do not have identities the way human faces do. \\
The final loss for the latent mapper ends up as follows:
\begin{equation}
    \begin{aligned}
        \mathcal{L}(V, w) = \:&H(0, C(V, G(w+M(w)))) \\
        &+ \lambda_{1}\, \lvert\lvert M(w) \rvert\rvert_2\\
        &+ \lambda_{2}\, \lvert\lvert G(w+M(w)) - V \rvert\rvert_2,
    \end{aligned}
\end{equation}
where $G$ is the pretrained generator, $C$ the pretrained comparator, and $V$ the source component of the inverted latent $w$.
\\Similarly, the optimization problem for the latent optimization is redefined as follows: \vspace{0.4cm}
\begin{equation}
    \lineskiplimit=-\maxdimen
    \begin{aligned}
        \argminA_{w \in \mathcal{W}} \:&H(0, C(V, G(w))) \\ 
        &+ \lambda_{1}\, \lvert\lvert w - w_s \rvert\rvert_2\\
        &+ \lambda_{2}\, \lvert\lvert G(w) - V \rvert\rvert_2,
    \end{aligned}
\label{equ:similarity_loss}
\end{equation}
where $w_s$ is the source latent code of the source component.

\section{Results and Discussion}
To test our optimization methods, we created a small toy dataset centered around the \textit{grabability} of components. The grabability of a component generally describes how well an industrial robot can grab onto it, which depends on the geometrical structure of the component, especially on the amount of even surface. For simplification, we reduce grabability to how well an industrial gripper can grab onto components, and the components were reduced to a specific class: screws and bolts. The dataset used for the training of the comparator contains 1000 pairs of screws/bolts with labels indicating which component in each pair has better grabability.
\subsection{Training and Evaluation of the StyleGAN}
The StyleGAN needs to be able to encode the components we want to optimize, which is why we trained the StyleGAN on a subset of the MCB dataset \cite{mcb}, filtered for classes visually similar to screws and bolts. This resulted in nine classes with a total of 22 788 components. Given this low amount of data, we deployed Adaptive Pseudo Augmentation \cite{apa} as limited data training strategy.\\

\begin{figure}[t!]
    \captionsetup{justification=centering}
    \centering
    \begin{subfigure}{.24\linewidth}
      \centering
      \includegraphics[width=\linewidth, trim=90 90 90 90, clip, frame, valign=m]{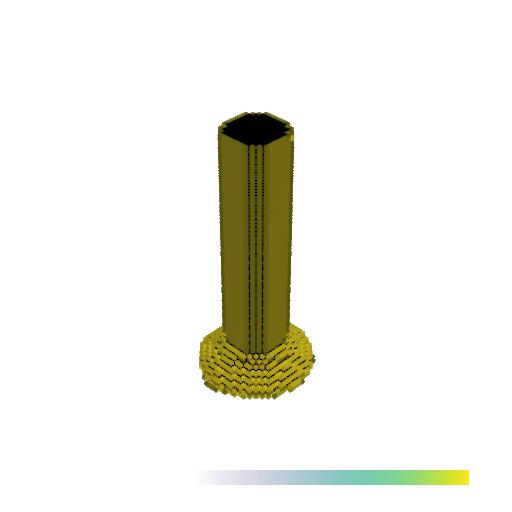}
    \end{subfigure}%
    \begin{subfigure}{.24\linewidth}
      \centering
      \includegraphics[width=\linewidth, trim=85 85 85 85, clip, frame, valign=m]{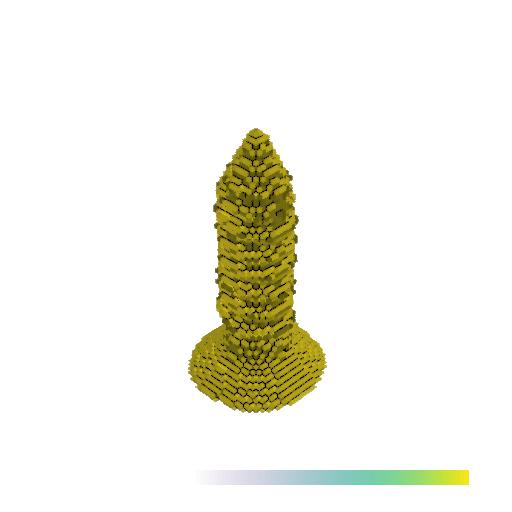}
    \end{subfigure}%
    \begin{subfigure}{.24\linewidth}
      \centering
      \includegraphics[width=\linewidth, trim=75 75 75 75, clip, frame, valign=m]{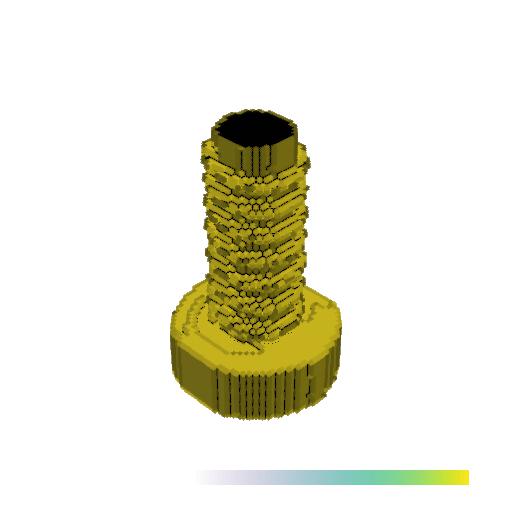}
    \end{subfigure}%
    \begin{subfigure}{.24\linewidth}
      \centering
      \includegraphics[width=\linewidth, trim=90 90 90 90, clip, frame, valign=m]{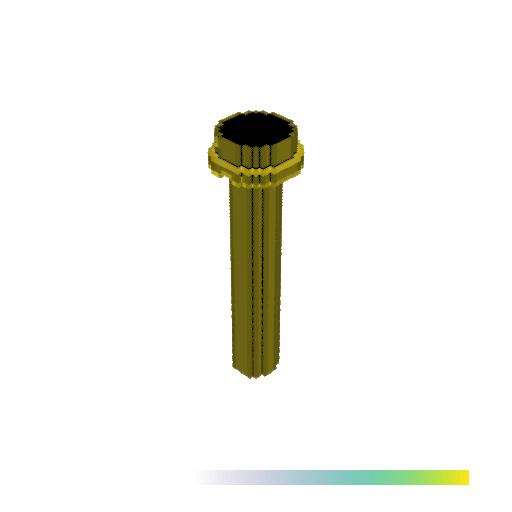}
    \end{subfigure}
        \begin{subfigure}{.24\linewidth}
      \centering
      \includegraphics[width=\linewidth, trim=80 80 80 80, clip, frame, valign=m]{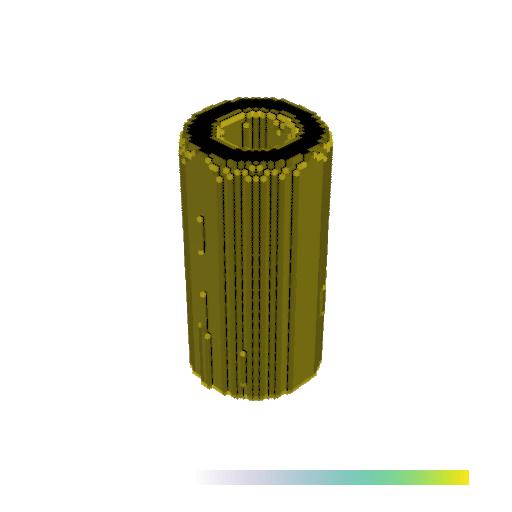}
    \end{subfigure}%
    \begin{subfigure}{.24\linewidth}
      \centering
      \includegraphics[width=\linewidth, trim=90 90 90 90, clip, frame, valign=m]{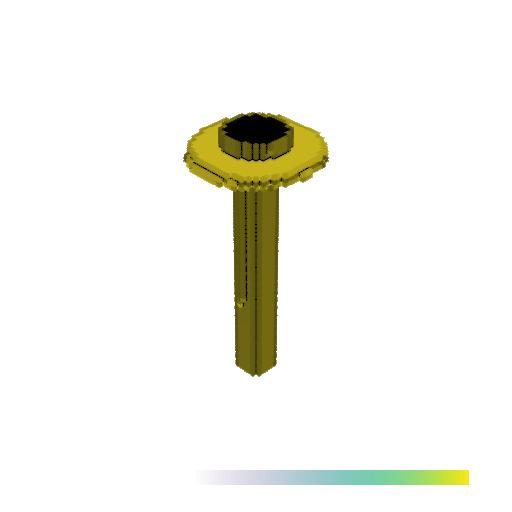}
    \end{subfigure}%
    \begin{subfigure}{.24\linewidth}
      \centering
      \includegraphics[width=\linewidth, trim=80 80 80 80, clip, frame, valign=m]{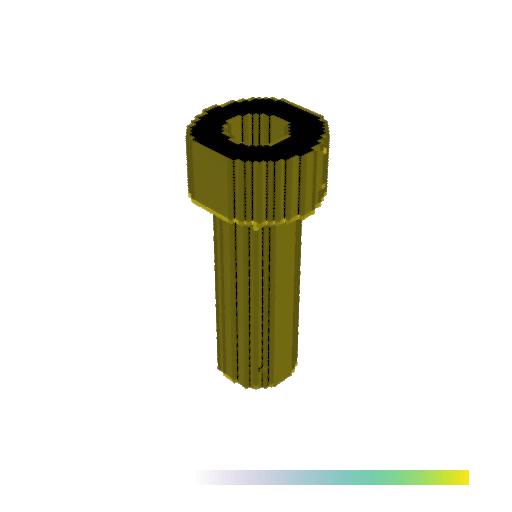}
    \end{subfigure}%
    \begin{subfigure}{.24\linewidth}
      \centering
      \includegraphics[width=\linewidth, trim=90 90 90 90, clip, frame, valign=m]{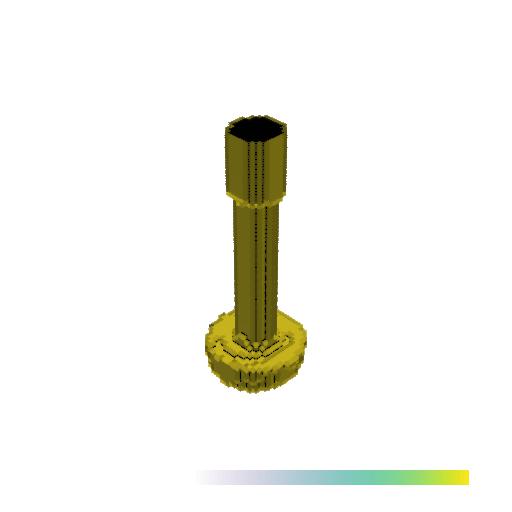}
    \end{subfigure}
        \begin{subfigure}{.24\linewidth}
      \centering
      \includegraphics[width=\linewidth, trim=95 95 95 95, clip, frame, valign=m]{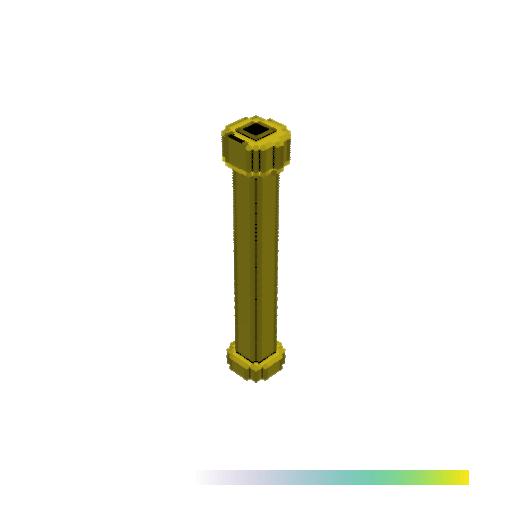}
    \end{subfigure}%
    \begin{subfigure}{.24\linewidth}
      \centering
      \includegraphics[width=\linewidth, trim=80 80 80 80, clip, frame, valign=m]{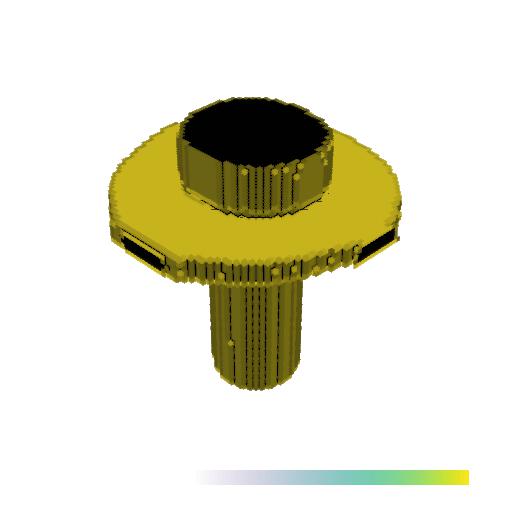}
    \end{subfigure}%
    \begin{subfigure}{.24\linewidth}
      \centering
      \includegraphics[width=\linewidth, trim=80 80 80 80, clip, frame, valign=m]{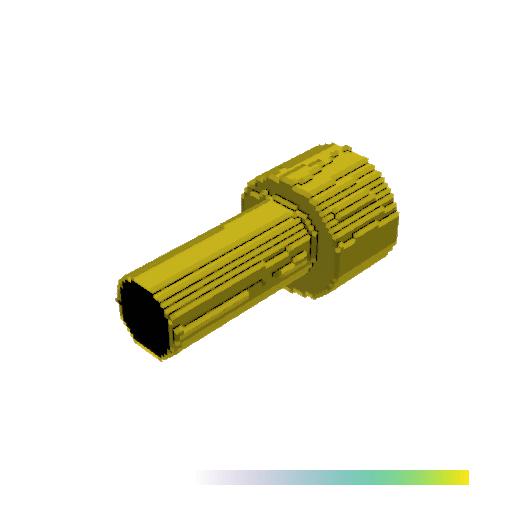}
    \end{subfigure}%
    \begin{subfigure}{.24\linewidth}
      \centering
      \includegraphics[width=\linewidth, trim=90 90 90 90, clip, frame, valign=m]{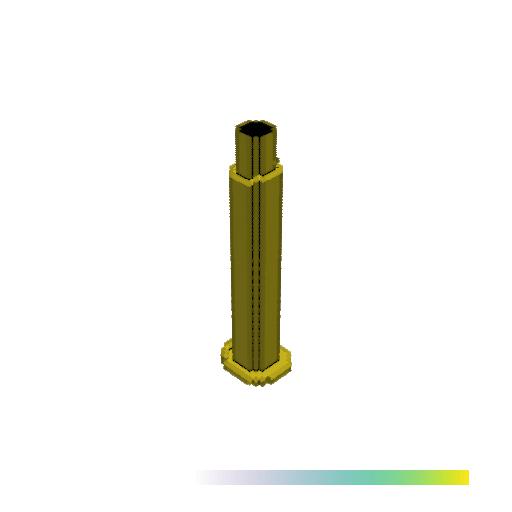}
    \end{subfigure}
        \begin{subfigure}{.24\linewidth}
      \centering
      \includegraphics[width=\linewidth, trim=90 90 90 90, clip, frame, valign=m]{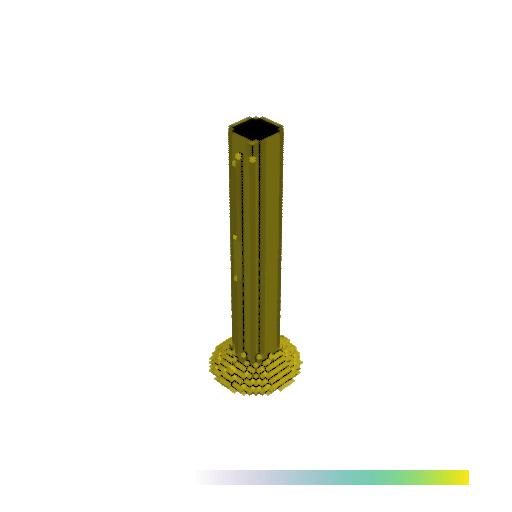}
    \end{subfigure}%
    \begin{subfigure}{.24\linewidth}
      \centering
      \includegraphics[width=\linewidth, trim=80 80 80 80, clip, frame, valign=m]{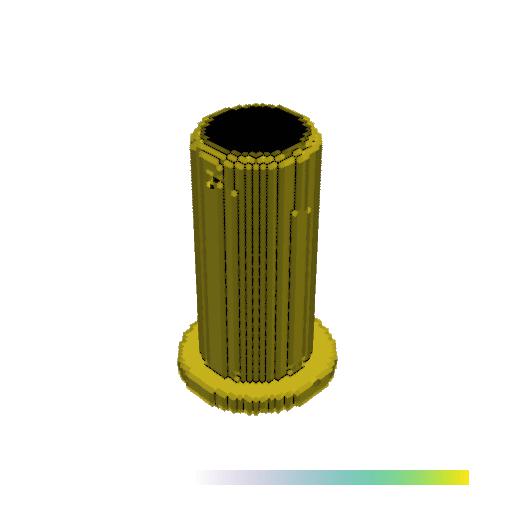}
    \end{subfigure}%
    \begin{subfigure}{.24\linewidth}
      \centering
      \includegraphics[width=\linewidth, trim=90 90 90 90, clip, frame, valign=m]{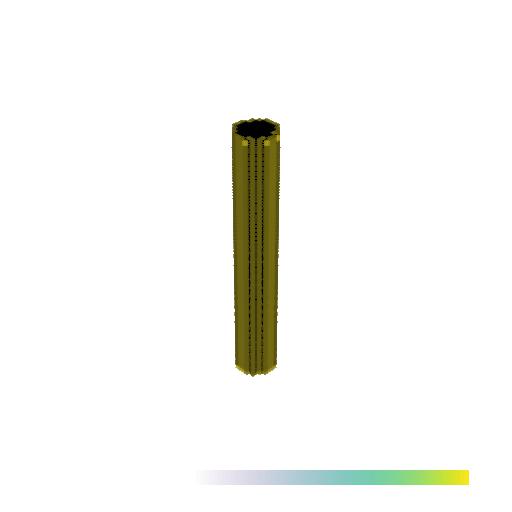}
    \end{subfigure}%
    \begin{subfigure}{.24\linewidth}
      \centering
      \includegraphics[width=\linewidth, trim=90 90 90 90, clip, frame, valign=m]{stylegan/step-0.jpg}
    \end{subfigure}%

    \caption{16 random components synthesized from the converged StyleGAN generator after
             training on the MCB dataset.}
    \label{stylegan_results}
\end{figure}

For the evaluation of the StyleGAN, the slice-wise FIDs were computed between all 22 788 images of the training dataset and 45 576 synthesized images, two synthesized images for each label in the training dataset. Since the InceptionV3 network cannot process the 64x64x1 middle slices directly, they were transformed to 128x128x3 through upsampling and replicating the channel dimension before being fed to the network.\\
We trained the StyleGAN until the slice-wise FIDs converged. This took about two days and the coronal, axial and sagittal FIDs converged to 90.77, 76.12 and 61.89 respectively. Figure \ref{stylegan_results} shows examples of synthesized components. 
\\We also attempted to train the StyleGAN on the ABC dataset \cite{abc}, which is a more complex and significantly larger dataset with over one million CAD models. Despite trying various training parameters and architecture modifications, the training did not succeed. It is unclear whether our setup was inappropriate or StyleGAN is not suited for training on more general and complex CAD datasets. \\

\subsection{Grabability Optimizations}

\begin{figure}[t!]
    \captionsetup{justification=centering}
    \centering
    \begin{subfigure}{.32\linewidth}
      \centering
      \includegraphics[width=\linewidth, trim=175 175 175 175, clip, frame, valign=m]{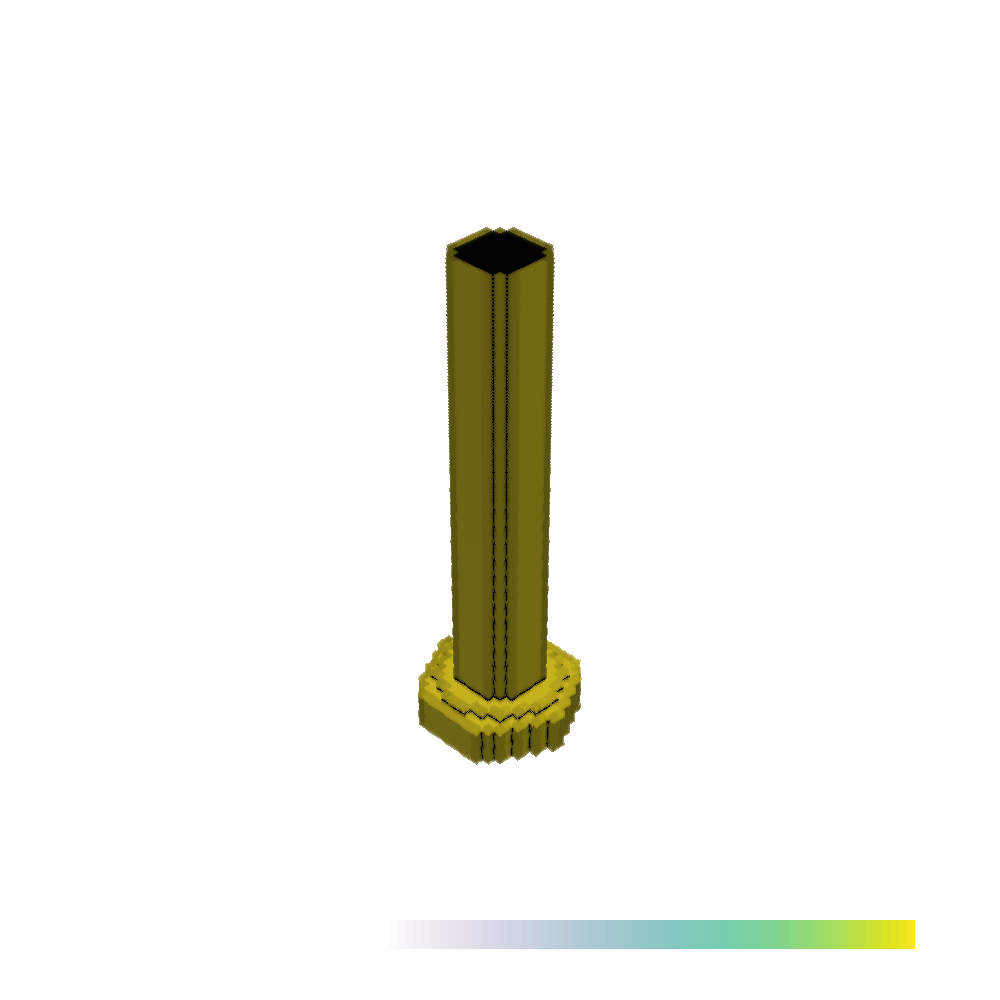}
    \end{subfigure}%
    \begin{subfigure}{.32\linewidth}
      \centering
      \includegraphics[width=\linewidth, trim=175 175 175 175, clip, frame, valign=m]{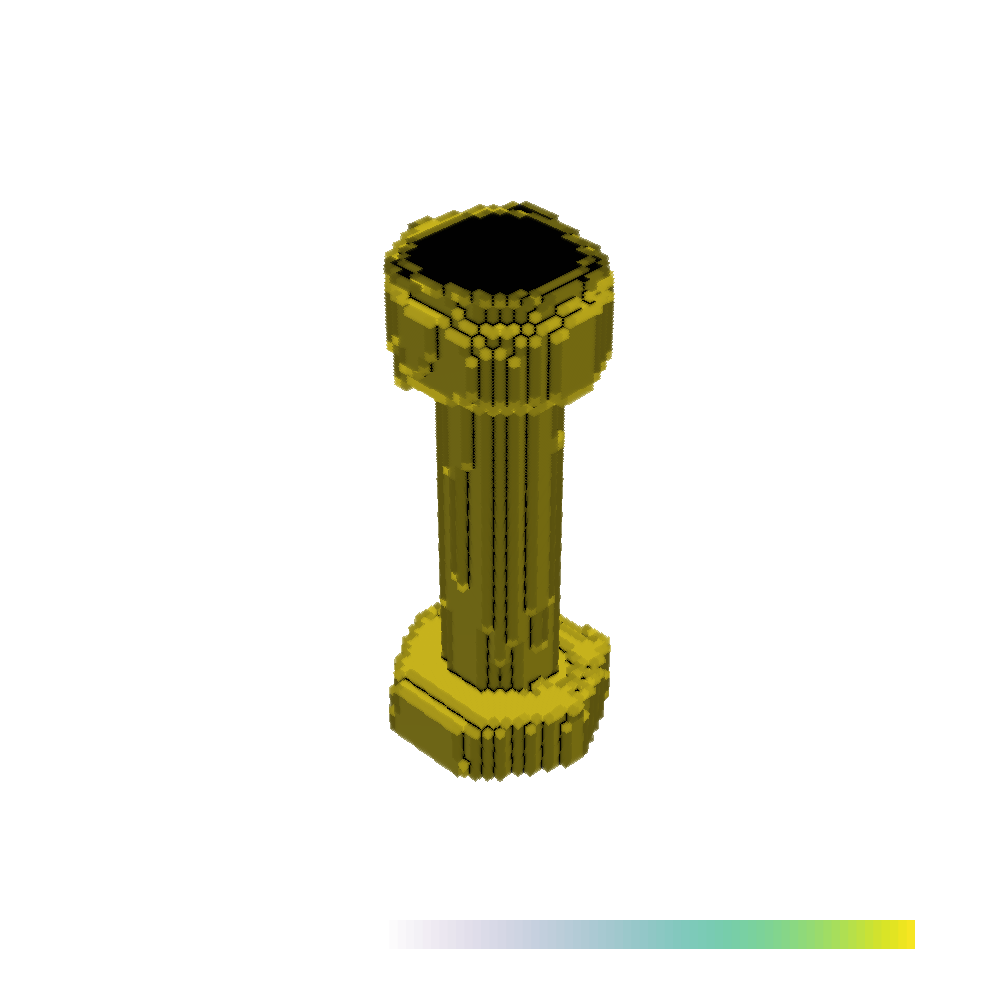}
    \end{subfigure}%
    \begin{subfigure}{.32\linewidth}
      \centering
      \includegraphics[width=\linewidth, trim=175 175 175 175, clip, frame, valign=m]{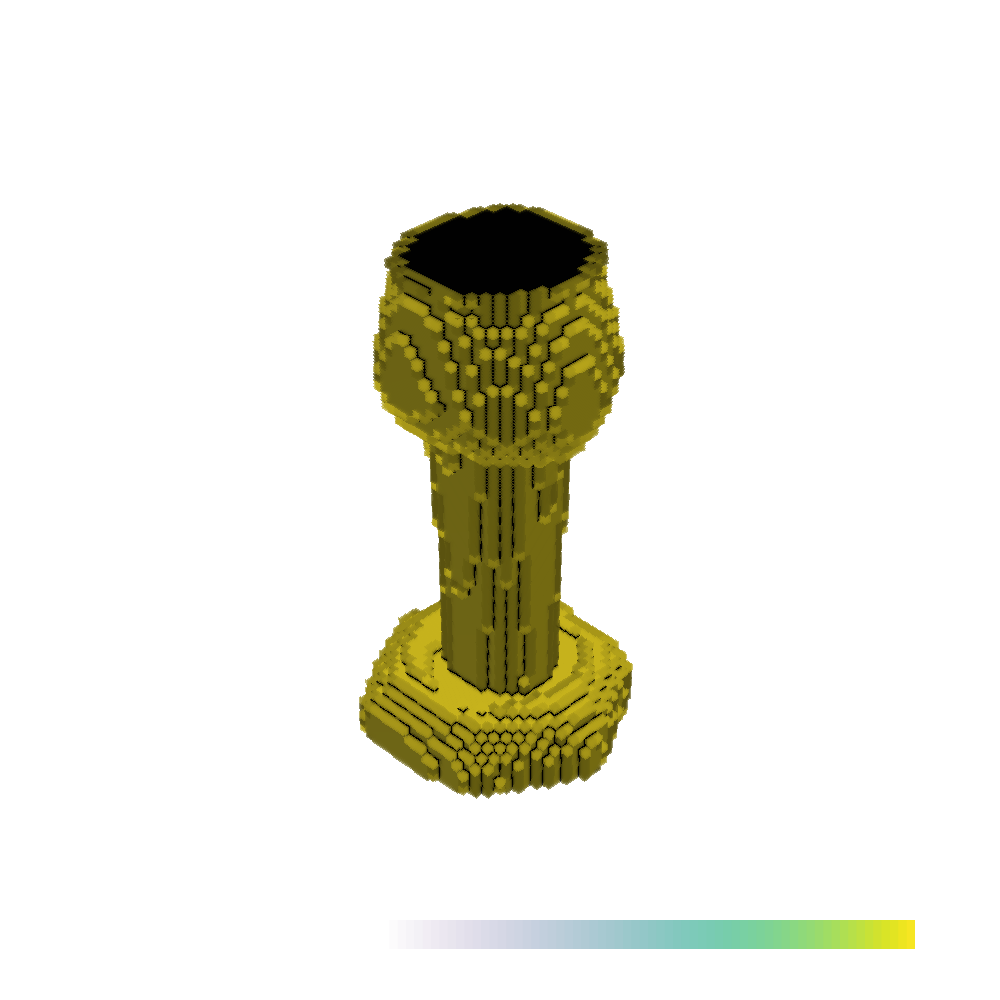}
    \end{subfigure} \\ \vspace{0.1cm}
    \begin{subfigure}{.32\linewidth}
      \centering
      \includegraphics[width=\linewidth, trim=150 150 150 150, clip, frame, valign=m]{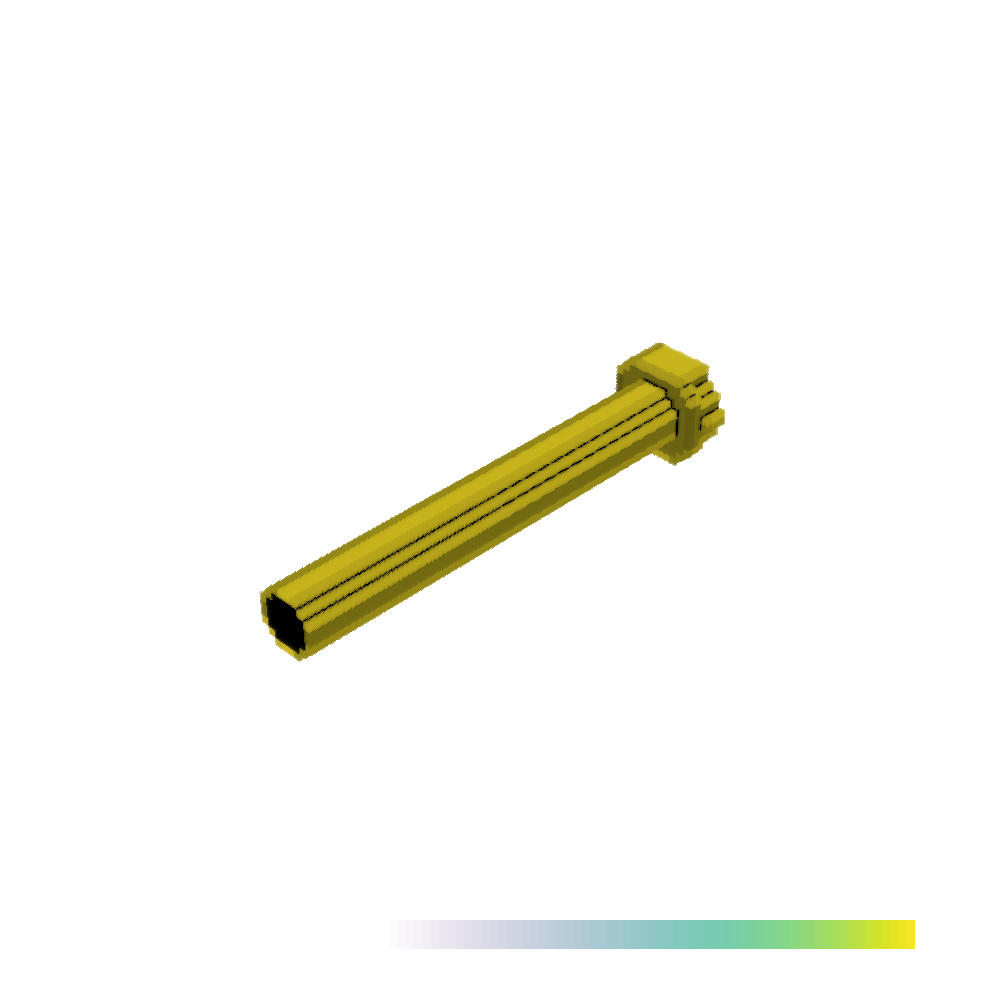}
    \end{subfigure}%
    \begin{subfigure}{.32\linewidth}
      \centering
      \includegraphics[width=\textwidth, trim=150 150 150 150, clip, frame, valign=m]{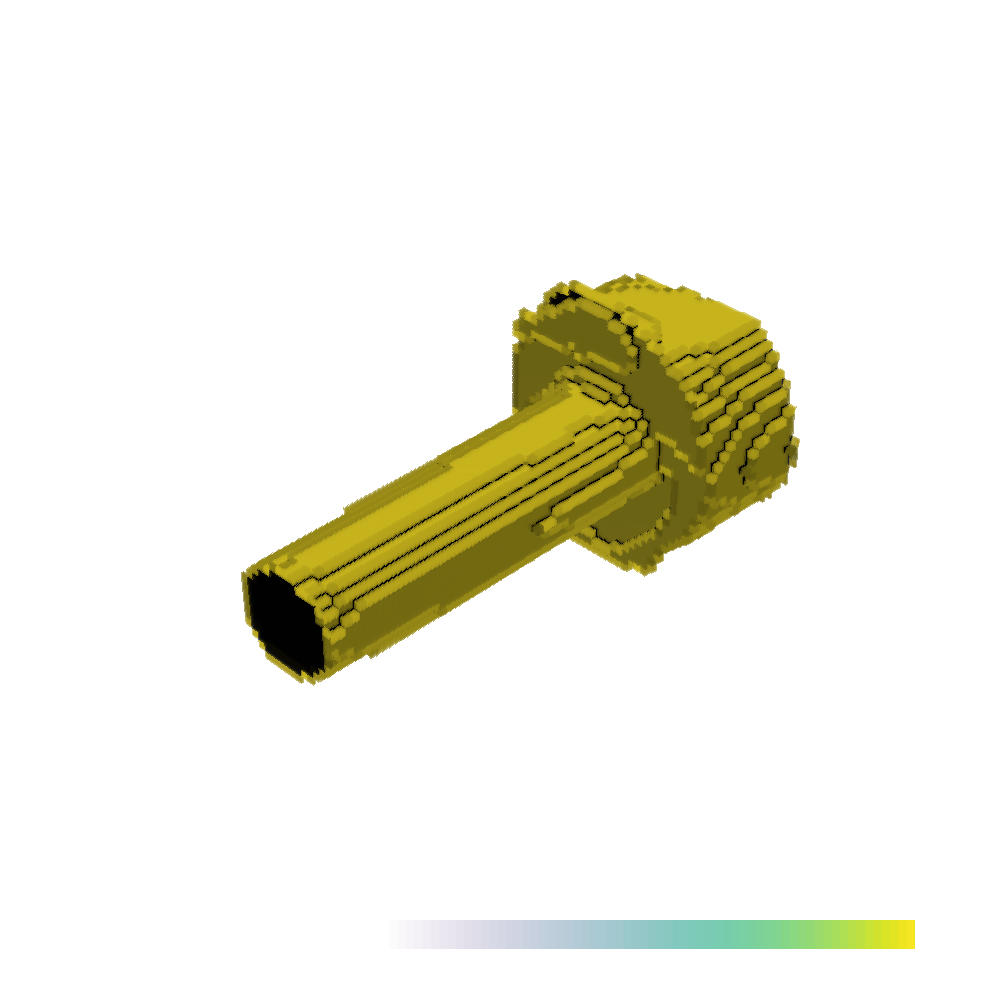}
    \end{subfigure}%
    \begin{subfigure}{.32\linewidth}
      \centering
      \includegraphics[width=\linewidth, trim=150 150 150 150, clip, frame, valign=m]{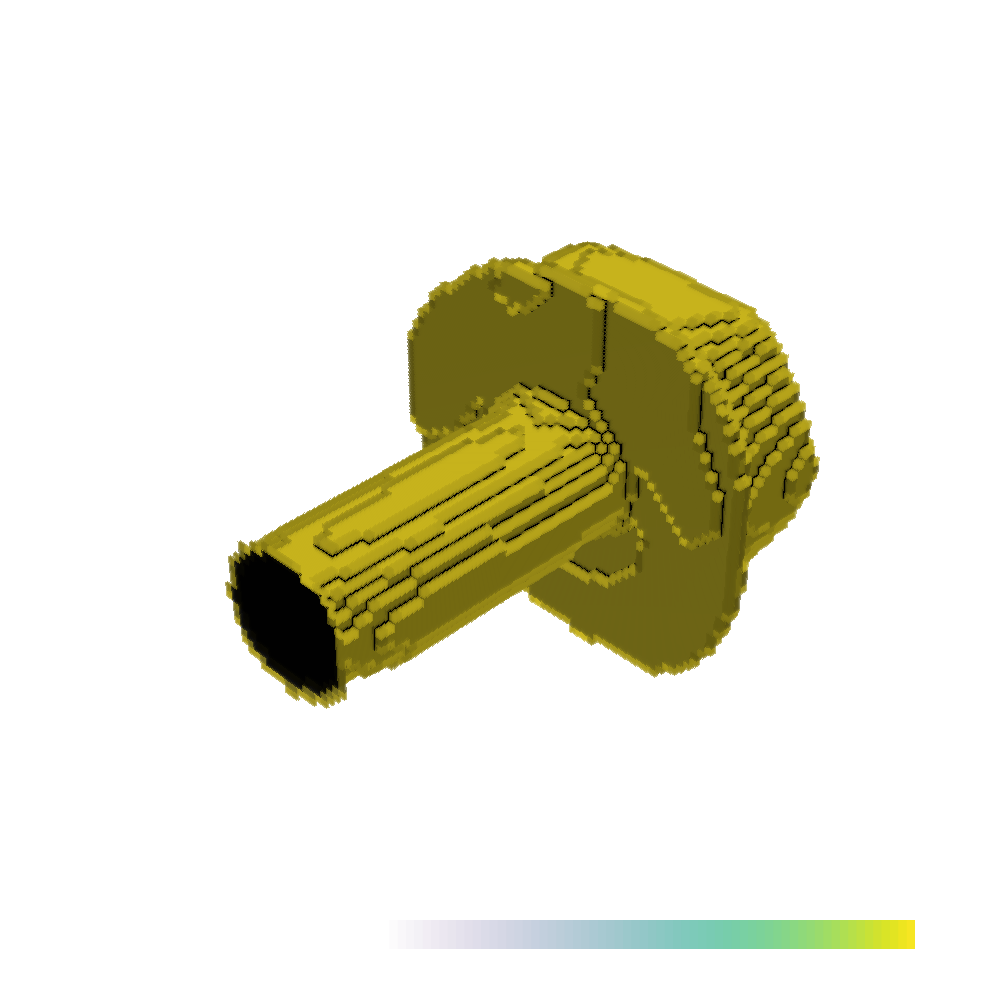}
    \end{subfigure} \\ \vspace{0.1cm}
    \begin{subfigure}{.32\linewidth}
      \centering
      \includegraphics[width=\linewidth, trim=175 175 175 175, clip, frame, valign=m]{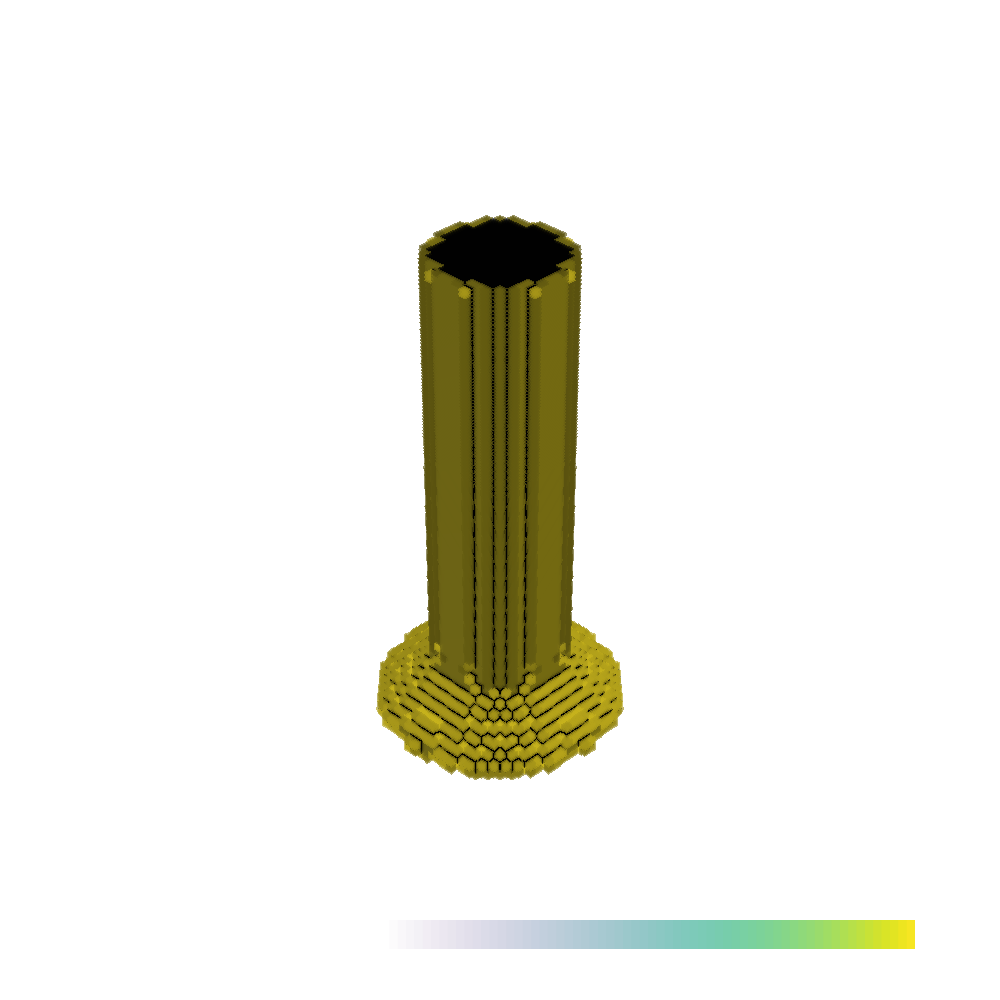}
    \end{subfigure}%
    \begin{subfigure}{.32\linewidth}
      \centering
      \includegraphics[width=\textwidth, trim=175 175 175 175, clip, frame, valign=m]{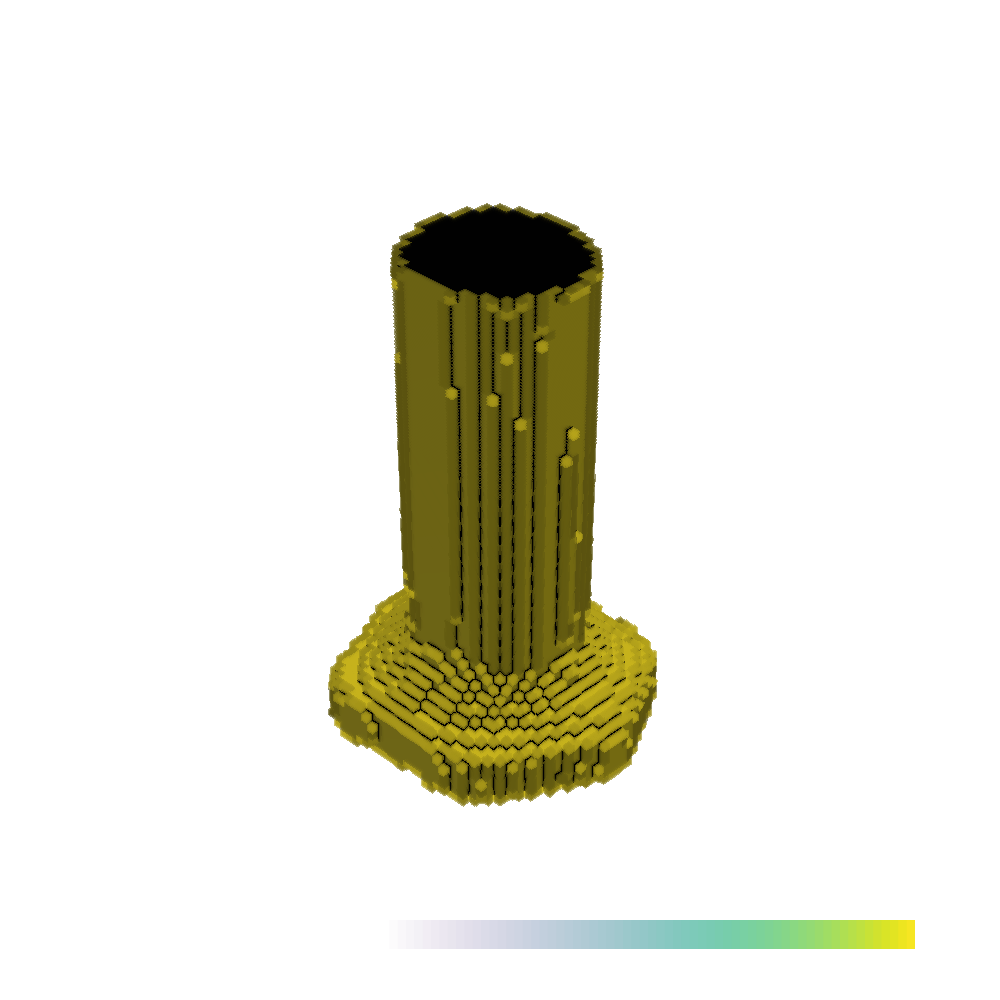}
    \end{subfigure}%
    \begin{subfigure}{.32\linewidth}
      \centering
      \includegraphics[width=\linewidth, trim=175 175 175 175, clip, frame, valign=m]{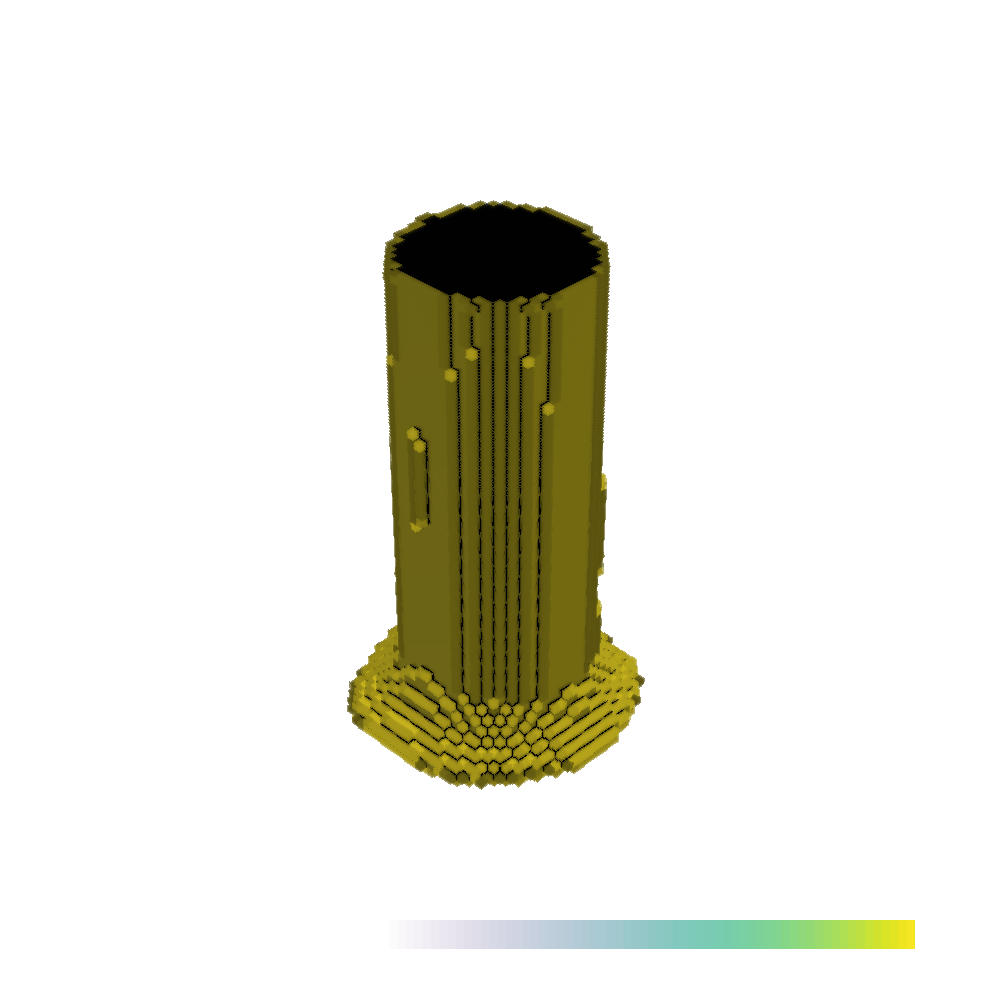}
    \end{subfigure} \\ \vspace{0.1cm}
    \begin{subfigure}{.32\linewidth}
      \centering
      \includegraphics[width=\linewidth, trim=175 175 175 175, clip, frame, valign=m]{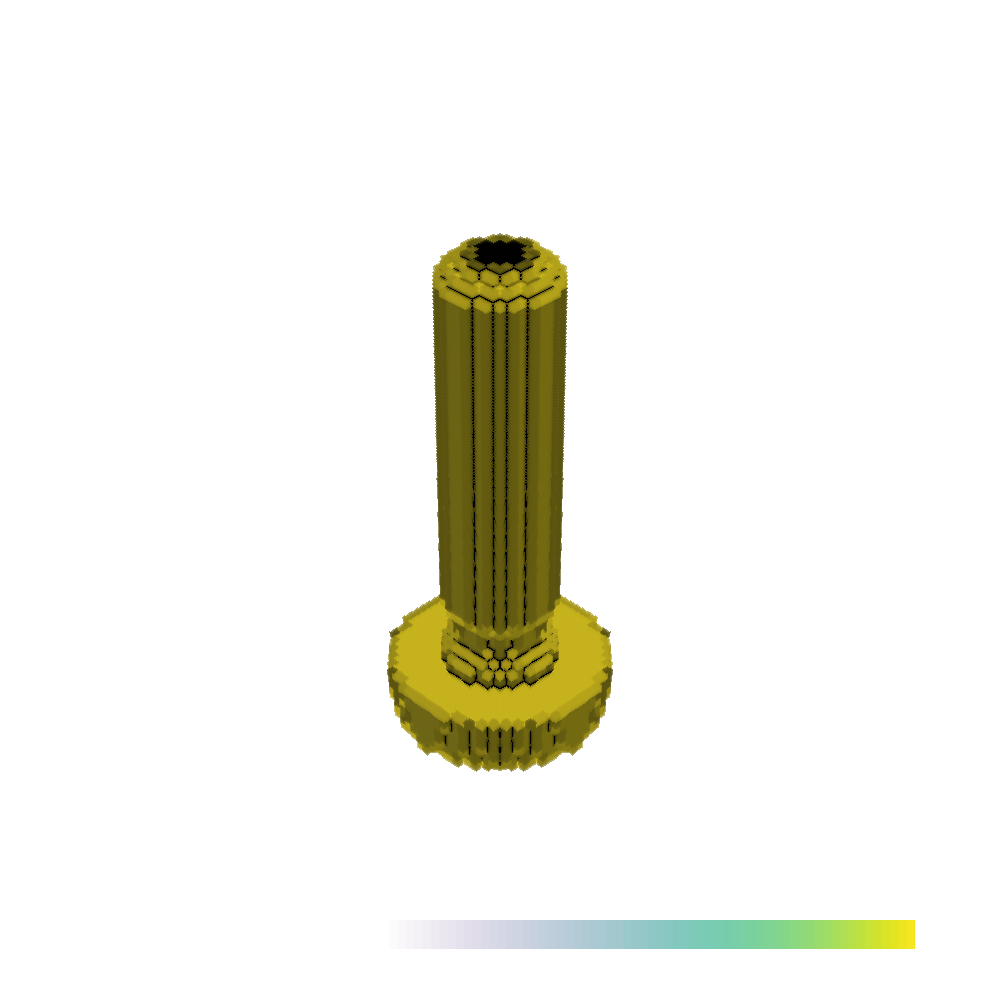}
      \caption{Original Component \newline}
    \end{subfigure}%
    \begin{subfigure}{.32\linewidth}
      \centering
      \includegraphics[width=\linewidth, trim=175 175 175 175, clip, frame, valign=m]{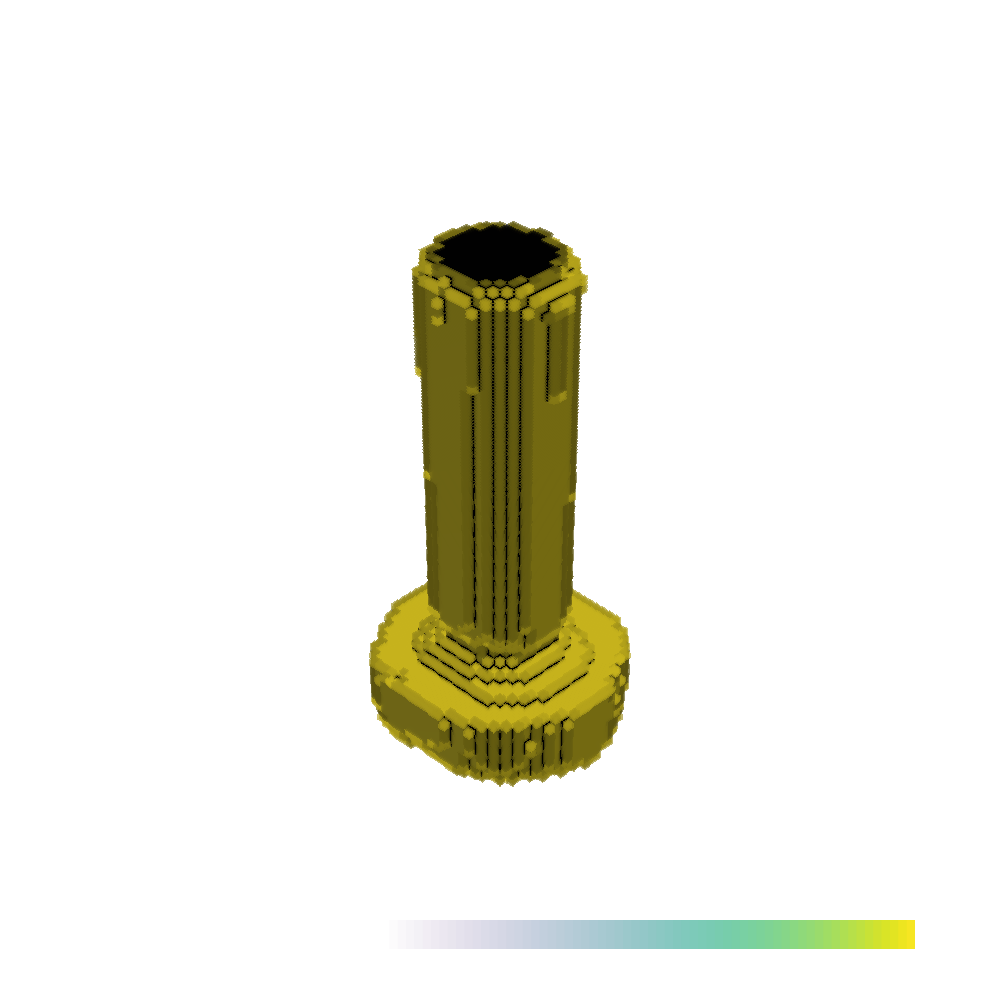}
      \caption{Latent Mapper \newline}
    \end{subfigure}%
    \begin{subfigure}{.32\linewidth}
      \centering
      \includegraphics[width=\linewidth, trim=175 175 175 175, clip, frame, valign=m]{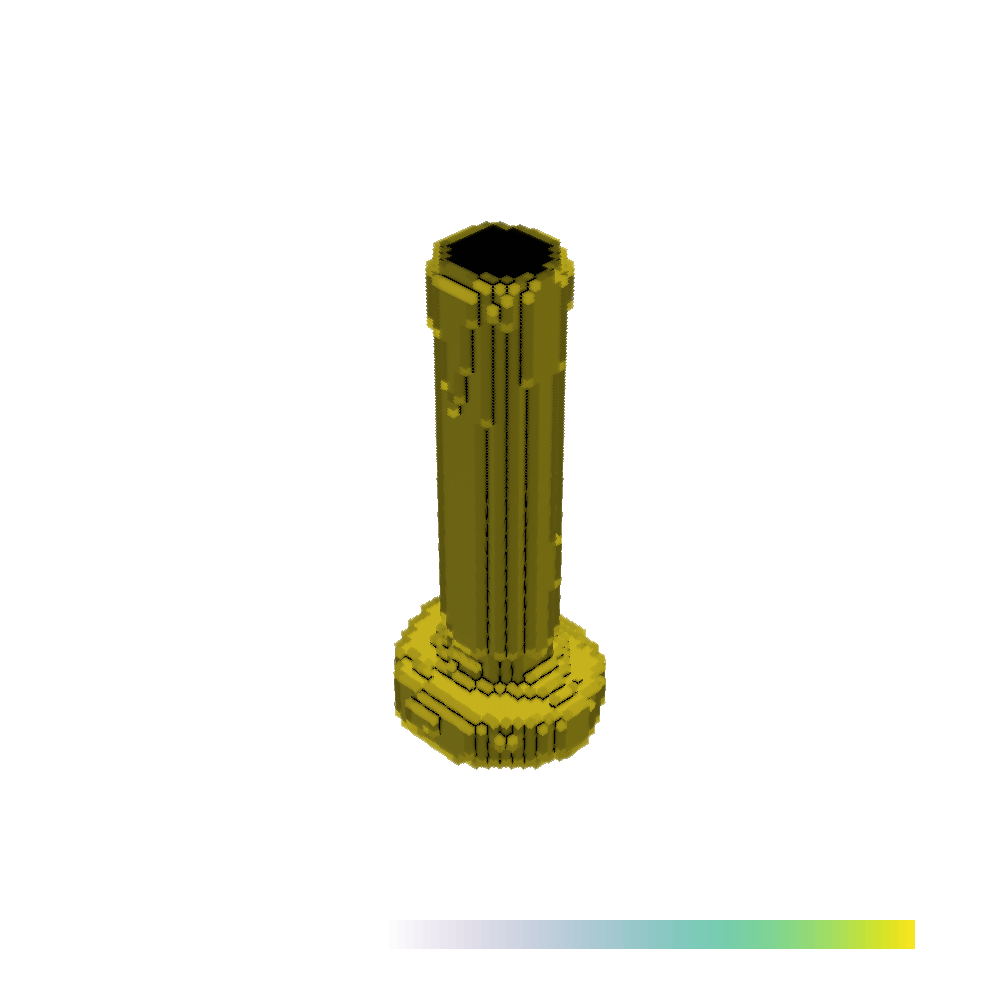}
      \caption{Latent \\ Optimization}
    \end{subfigure}
    \caption{Grabability optimizations using the latent mapper and latent optimization.}
    \label{grabability_optimizations}
\end{figure}

\begin{figure}[t!]
    \centering
    \begin{subfigure}{.24\linewidth}
      \centering
      \includegraphics[width=\linewidth, trim=175 175 175 175, clip, frame, valign=m]{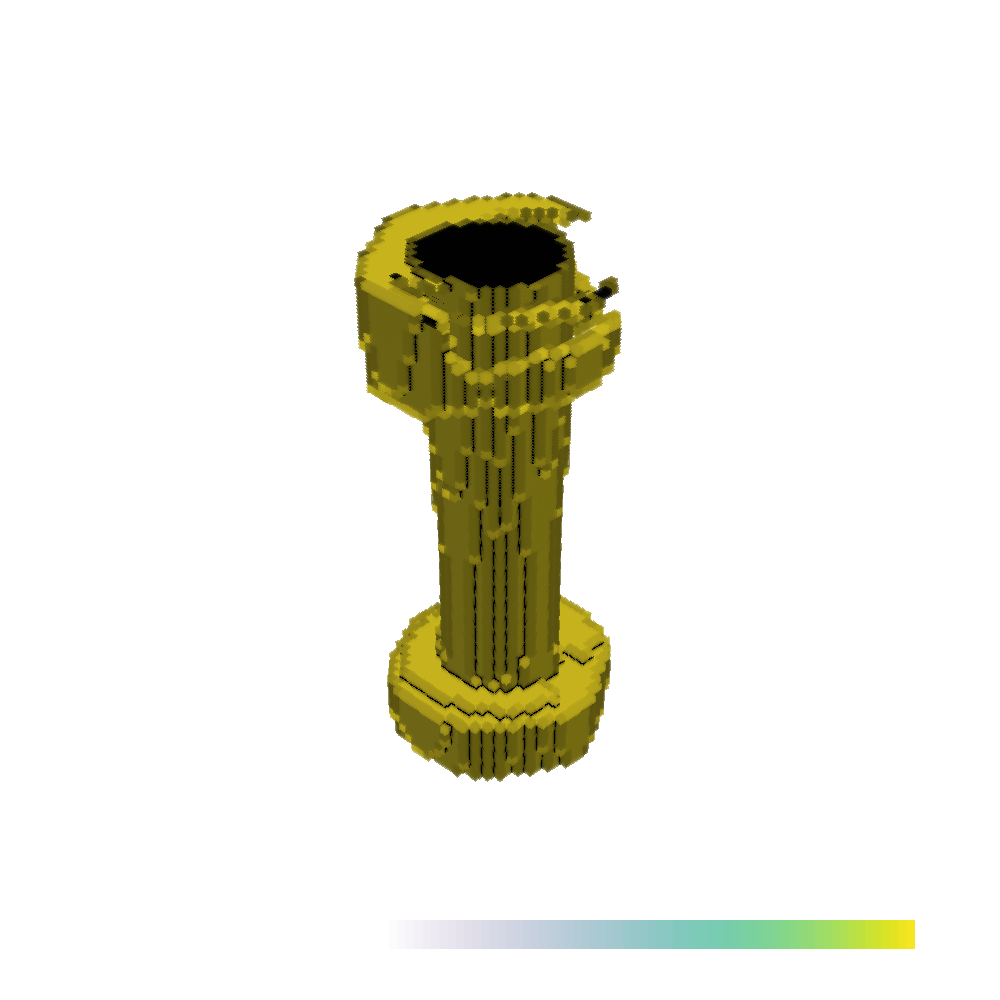}
    \end{subfigure}%
    \begin{subfigure}{.24\linewidth}
      \centering
      \includegraphics[width=\linewidth, trim=100 100 100 100, clip, frame, valign=m]{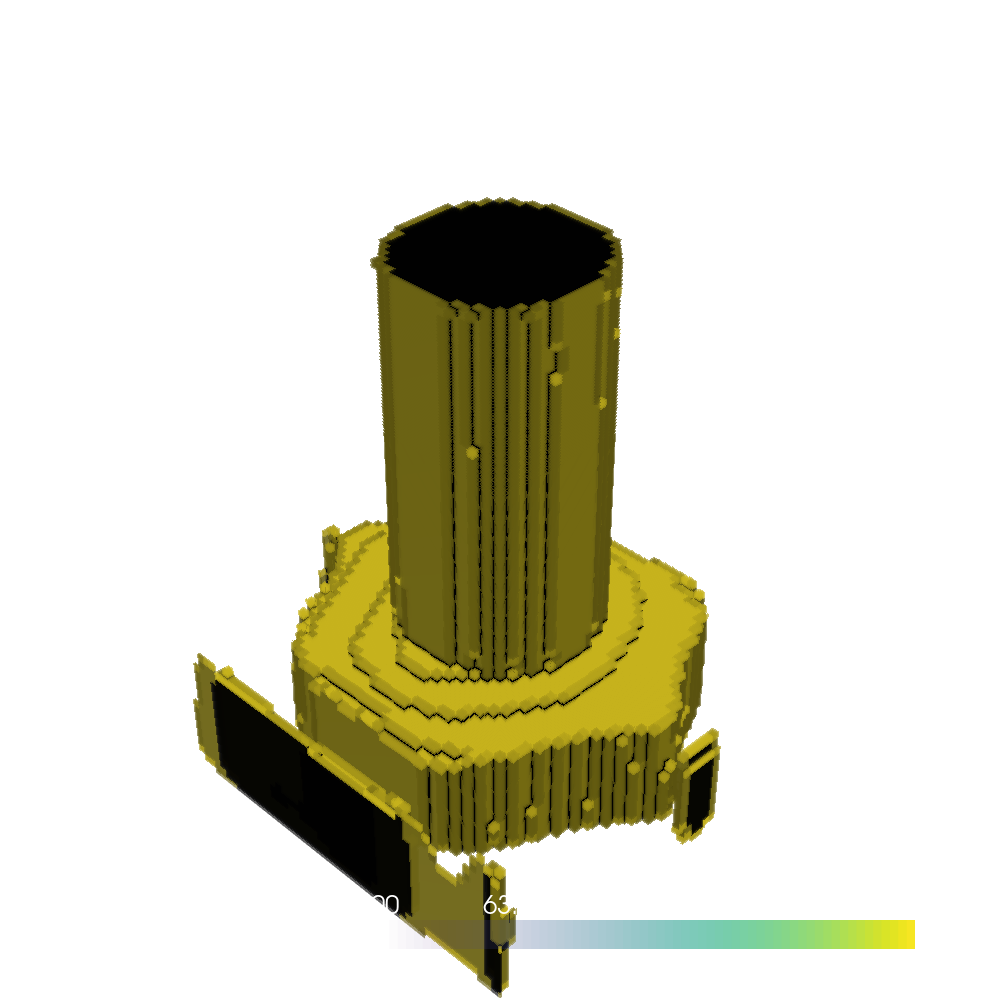}
    \end{subfigure}%
    \begin{subfigure}{.24\linewidth}
      \centering
      \includegraphics[width=\linewidth, trim=175 175 175 175, clip, frame, valign=m]{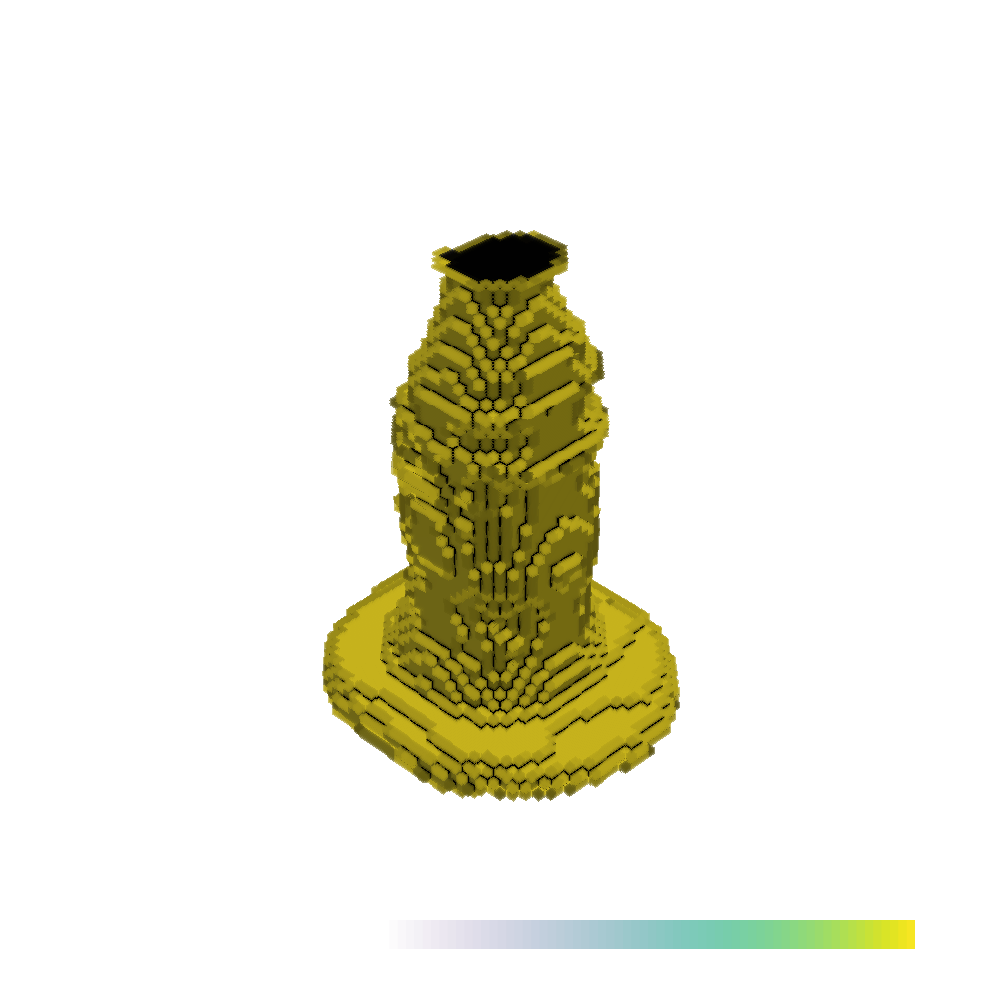}
    \end{subfigure}%
    \begin{subfigure}{.24\linewidth}
      \centering
      \includegraphics[width=\linewidth, trim=100 100 100 100, clip, frame, valign=m]{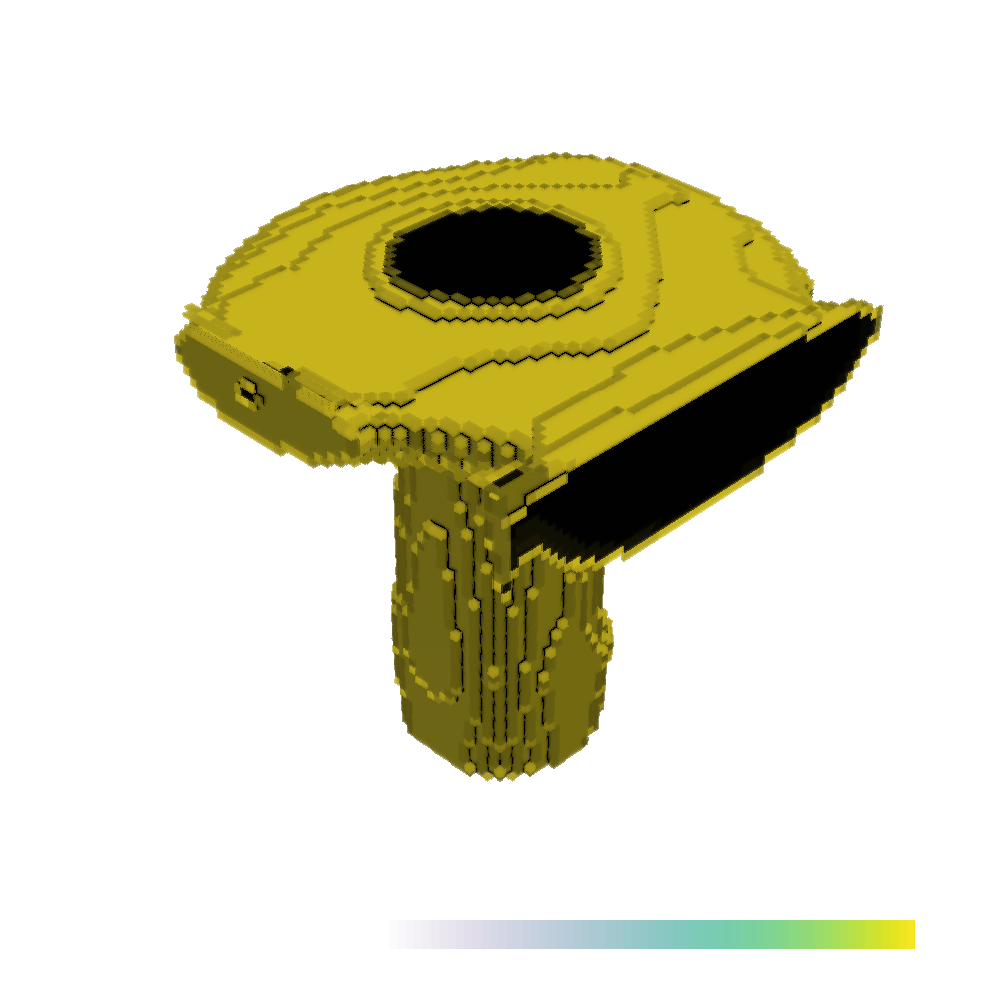}
    \end{subfigure}
    \caption{Examples of unrealistic or distorted optimizations}
    \label{problematic_optimizations}
\end{figure}

After the training of the StyleGAN and the subsequent training of the latent mapper, we ran the optimization methods with various parameters. In Figure \ref{grabability_optimizations}, we provided several examples of grabability optimizations we handpicked for their perceptual quality and reasonableness. Grabability is mostly modeled by how much even surface there is to grab onto, and the optimizated components mainly increase the size of the screws' and bolts' heads, which aligns with our grabability model. Unfortunately, many of the optimizations are more akin to the ones in Figure \ref{problematic_optimizations}, where they have obvious flaws such as appearing incomplete and distorted, or having significant artifacts, i.e. generally being unrealistic.\\
The unrealistic optimizations make practical applications difficult, and also negatively affect the training of the latent mapper, during which they may confuse the comparator given that it is only trained on realistic images.\\
Current latent space manipulation methods have no incentive for realistic results, and instead rely on the properties of the latent space. Particularly, they rely on the latent codes in a close area around the source latent code also resulting in mostly realistic images.
For the manipulation of CAD models, this is insufficient, since CAD models typically follow stricter rules regarding their shape, which manipulations in latent space are unlikely to respect.\\
The problem therefore stems from the data representation and is not unique to voxel models, but also encompasses other popular representations such as point clouds or view-based representations. Future work could explore using generative models that use natural language representations where CAD models are defined with the operations used to create them \cite{deepcad}. Compared to our 3D StyleGAN, GAN models that use natural language to represent CAD models have also been shown to work with significantly more complex datasets while substantially reducing training costs.\\
Lastly, it is often desirable to protect specific structures of the component, for example if they serve a core functionality. The latent optimization method can be extended to allow for this by adding an additional $L_2$ loss that punishes changes to protected structures, or by reversing changes to protected structures in G(w). Extending the latent mapper is more challenging, since it would need a mechanism by which to select reasonable protected structures during training. \\

All code we used for our experiments can be found in our repository \cite{GitHub}. \\

\section{Conclusion}

Within this work, we presented how image manipulation methods such as StyleCLIP can be adapted for the optimization of CAD models. We also demonstrated the ability of our system by using it to optimize the grabability of screws and bolts. Many of the optimizations have high perceptual quality and are reasonable in the context of our definition of grabability, but there are still obvious flaws such as unrealistic or distorted optimizations that plague the results.
Considering our very limited dataset of 1000 samples and the fact that changing from voxel representations to natural language representations promises to fix the identified flaws of our system, we nevertheless see great potential in using latent space manipulation for the optimization of abstract CAD features. We think that systems like ours can improve the automation of manufacturing by creating an alternative to needing the knowledge and experience of domain experts for the design of components fit for automation processes.

\section{Future Work}

The optimization methods both have several limitations. The quality of the optimized models is not consistent in symmetry and functionality. Sometimes, the wrong feature is optimized or the optimization is distorted. In addition, the training of 3D-CNNs is very computing-intensive. From the usability perspective, there is also a need to identify and mark certain functional structures as protected. Tackling these challenges will be necessary in order to achieve an industry ready optimization algorithm.


\addtolength{\textheight}{-12cm}   




\section*{Acknowledgments}

The researchers would like to thank the German Federal Ministry for Education and Research (BMBF) as well as the Projektträger Karlsruhe (PTKA) for the funding within the project “KiMont” (Grant number 02K20K512).

\clearpage\onecolumn

\end{document}